\documentclass{article}
\usepackage{ijcai23}

\usepackage{times}
\usepackage{soul}
\usepackage{url}
\usepackage[hidelinks]{hyperref}
\usepackage[utf8]{inputenc}
\usepackage[small]{caption}
\usepackage{graphicx}
\usepackage{amsmath}
\usepackage{amsthm,amssymb,bm}
\usepackage{booktabs}
\usepackage{algorithm}
\usepackage{algorithmic}
\usepackage{subcaption}
\usepackage{epstopdf}
\usepackage[table]{xcolor}
\usepackage{multirow}
\usepackage[switch]{lineno}
\usepackage{makecell}
\usepackage{color}

\title{Multi-view Semantic Consistency based Information Bottleneck for Clustering}

\author{
Wenbiao Yan$^1$
\and
Jihua Zhu$^1$\and
Yiyang Zhou$^{1}$\And
Yifei Wangr$^1$
Qinghai Zheng$^2$
\affiliations
$^1$School of Software Engineering, Xi'an Jiaotong University, Xi'an 710049, China\\
$^2$College of Computer and Data Science, Fuzhou University, Fuzhou 350108, China\\
\emails
wenbiao777@stu.xjtu.edu.cn
}

\begin{document}

\maketitle
	
\begin{abstract}
	Multi-view clustering can make use of multi-source information for unsupervised clustering. Most existing methods focus on learning a fused representation matrix, while ignoring the influence of private information and noise. To address this limitation, we introduce a novel Multi-view Semantic Consistency based Information Bottleneck for clustering (MSCIB). Specifically, MSCIB pursues semantic consistency to improve the learning process of information bottleneck for different views. It conducts the alignment operation of multiple views in the semantic space and jointly achieves the valuable consistent information of multi-view data. In this way, the learned semantic consistency from multi-view data can improve the information bottleneck to more exactly distinguish the consistent information and learn a unified feature representation with more discriminative consistent information for clustering. Experiments on various types of multi-view datasets show that MSCIB achieves state-of-the-art performance.
\end{abstract}

\section{Introduction}

As an unsupervised task, cluster analysis is widely used in machine learning, computer vision, and data mining. Compared with cluster analysis using only single-view features, each sample can often be described from its different features. These descriptions come from different feature extractors, such as visual features, text features, and multiple mappings of the same object, which are derived from different hardware and extraction networks. For this heterogeneous information, multi-view clustering(MVC) aims to improve the effectiveness of clustering models by mining the consistent information hidden in different views, which has become an important research direction in unsupervised tasks.

Since there is a natural correspondence between different views in multi-view data, and for a multi-view sample, multiple views describe the same object, and there is common semantics in multi-view data. Multi-view clustering method mainly focuses on learning the consistent information of multi-view features so as to further learn the clustering information of samples. Existing MVC methods can be roughly divided into the following categories: subspace-based clustering methods, which explore the common representation of multiple views and learns the consistent information of samples. The other type of method is based on non-negative matrix decomposition. \cite{MVCJNMF} learns the low-rank matrix of each view for clustering, eliminating redundant information of view features. The third type is based on graph structure information, which is learned in a cluster of multiple views. \cite{ijcai2017p357,ijcaiGFMAGC} address this problem by exploring a Laplacian rank-constrained graph. Several neural network-based methods have been extensively developed recently.

\begin{figure*}[htbp]
	\centering
	\includegraphics[scale=1]{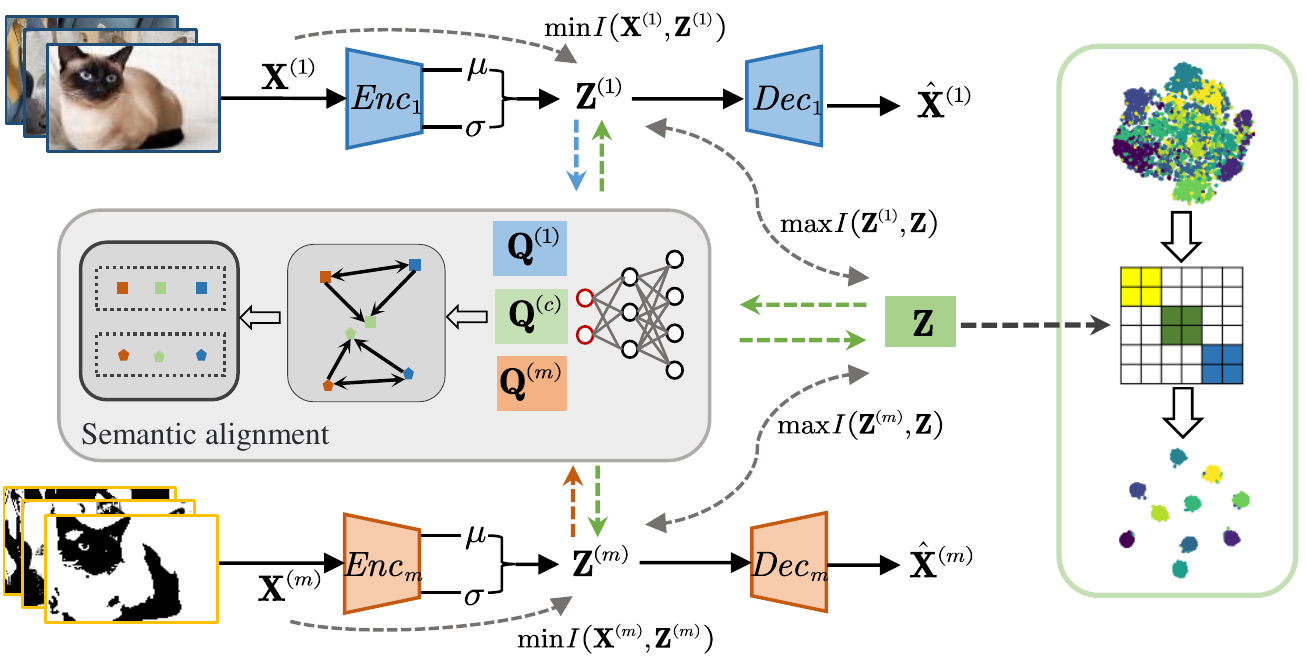}
	\caption{MSCIB framework, which learns a more discriminative consistent representation by utilizing the information bottleneck theory to reduce the adverse effects of task-irrelevant redundant information. Without loss of generality, two views are taken here as examples in the framework, where RGB and grayscale data are taken as input. The solid black arrow depicts the process of feature representation learning and the \textcolor{gray}{gray} dashed line indicates the part of the information bottleneck. The \textcolor{orange}{orange} and \textcolor{cyan}{cyan} parts demonstrate how the model processes the two views of the sample, and the \textcolor{green}{green} part denotes the representation to be learned. Specifically, $Enc_m$ denotes the variational autoencoder. The reconstruction loss is designed on the within-view representation $\mathbf{Z}^{(m)}$, and the semantic consistency loss is built on the semantic features $\mathbf{Q}$ learned from the feature representations $\mathbf{Z}^{(m)}$ with $\mathbf{Z}$.}
	\label{fig2}
\end{figure*}

Due to the powerful fitting ability of the neural network, the deep MVC method has more and more excellent presentation ability. Many MVC practices \cite{ijcai2019p409,EtEAANMMC} incorporate features of all views $\{ \mathbf{Z}^{(m)} \} _{m=1}^{M}$ to achieve a common representation of all views. The multi-view clustering task is transformed into a single-view clustering task by directly clustering the fused features. However, each view feature $\mathbf{X}^{(m)}$ contains common semantics and private information. The latter is meaningless, even misleading, and may interfere with the quality of the fusion features, resulting in poor clustering. Some MVC approaches \cite{iccv2019RMLSLMVC,cvpr2021COMPLETER} use consistent goals on underlying features to explore common semantics across all views. However, they usually require reconstruction of the objective to avoid trivial solutions. The consistency objective learns the features with common semantics in all views as much as possible, while the reconstruction objective desires to use the same features to keep the view private information of each view. However, it is difficult to distinguish common information from redundant information by only using the consistent objective and reconstruction objective in the feature space.

To address this contradiction, we propose a new framework named Multi-view Semantic Consistency based Information Bottleneck for clustering (MSCIB). This framework avoids direct feature fusion but establishes learnable feature representations and semantic embedding for each view. It utilizes information bottleneck theory to guide the extraction of consistent information and introduce the semantic consistency loss constraint, which can decrease the distance between similar view representations. Information bottleneck theory allows the learned representation to retain clustering information as much as possible but discard other information as much as possible. Therefore, MSCIB considers both contrastive consistency and reconstruction performance of feature representations. What’s more, it also pursues semantic consistency to improve the information bottleneck learning process.

The MSCIB framework is displayed in Figure \ref{fig2}. The goal of MSCIB is to remove the redundant information in samples and learn a consistent representation matrix Z for the clustering task. Specifically, MSCIB utilizes the variational autoencoder and reparameterization method to learn the within-view representation features of each view. For the model constructability, the learned representation features to reconstruct original features, where Information bottleneck theory is used to constrain the representation features. Meanwhile, the MLP network takes the learned representation features as input to learn the semantic features $\mathbf{Q}$, where consistent constraints are imposed by the contrastive learning to enhance their consistency. The main contributions of this paper is delivered as follows:
\begin{itemize}
	\item We propose a novel deep multi-view clustering method to mine the common information of multi-view features while remove the redundant information of original features, which improves the interpretability of feature representation.
	\item We use semantic consistency to guide the information bottleneck theory for learning more discriminative feature representation.
	\item We validate the effectiveness of MSCIB by conducting extensive experiments on several datasets. Experimental results illustrate that our MSCIB method has superior performance over several sate-of-arts MVC methods.
\end{itemize}

\section{Related Work}

In general, multi-view clustering methods can be divided into five categories. The first category is based on multi-view subspace clustering methods, which assumes that feature data from multiple views share a common subspace. \cite{9010687} explores a self-representation layer to hierarchically reconstruct view-specific subspaces and uses an encoding layer to make cross-view subspaces more consistent. The second category is based on non-negative matrix factorization, \cite{9170204} explores the common latent factors of multiple views through matrix factorization. The third category is based on Canonical Correlation Analysis. CCA was first proposed by \cite{Hotelling1992}, which processed paired datasets to find the linear transformations of each view that maximize the correlation between the transformed variables. The fourth category is graph-based methods, which first obtain a similarity matrix for each view. \cite{ijcai2017p357} assumes a common index matrix. The problem is then solved by minimizing the difference between the common indicator matrix and each similarity matrix. The fifth category is methods based on deep learning frameworks, \cite{AE2NET,SDMVC,MFLVC,ijcai2019p356} utilize autoencoders to capture the latent clustering features of multi-view data.

The information bottleneck(IB) theory provides an information theoretic approach to representation learning. \cite{alemi2017deep} uses neural networks to parameterize the information bottleneck model and exploits the reparameterization trick for effective training. The original information bottleneck formulation requires data with supervised information for learning. However, in unsupervised multi-view clustering tasks, the lack of labels limits the development of information bottleneck theory. To explore the nonlinear relationship of multiple views, a deep multi-view information bottleneck network is proposed in \cite{wang2019deep}. \cite{Federici2020Learning} extends the information bottleneck approach to the unsupervised multi-view setting. In the two-view task, the mutual information of features and representations is cross-computed to maximize the retention of prediction information and discard redundancy. Guided by information bottleneck theory, \cite{wan2021multi} extensively uses the common representation between each view and the specific representation of a single view to learn a feature representation with rich label information.

\section{The Proposed Method}

In this work, we propose MSCIB to learn a parsimonious and discriminative representation. Give a multi-view dataset $\{\mathbf{X}^{(m)}\in R^{N\times D_m}\}_{m=1}^{M}$, including $N$ instances across $M$ views, where $D_m$ denotes the dimensionality of the $m$-th view. Multi-view clustering aims to partition the examples into $K$ clusters.

The proposed framework is shown in Figure \ref{fig2}, its key components include information bottleneck and semantic consistency. Overall, the loss function of our method can be formulated as follows:
\begin{equation}
	\label{loss_all}
	\mathcal{L}=\mathcal{L}_{Rec}+\lambda _1\mathcal{L}_{IB}+\lambda _2\mathcal{L}_{Sem},
\end{equation}
where $\mathcal{L}_{Rec}$ is the reconstruction loss of each view, $\mathcal{L}_{IB}$ indicates the loss of information bottleneck theory, and $\mathcal{L}_{Sem}$ is the loss of semantic consistency. In this way, irrelevant information in the original view can be reduced, resulting in a compact and discriminative representation that encodes view-specific intrinsic information. The parameters $\lambda _1$ and $\lambda _2$ are used to trade off the effects of different losses.

\subsection{Feature Reconstruction}
Usually, the multi-view feature dimensions of a sample are different, and the multi-view data have redundancy and random noise. To reduce the influence of adverse information on the clustering task and compress the original features to obtain the representation features of the same dimension, the variational autoencoder is used to obtain the sample features for encoding. The independent view decoder is also used to recover the original part of the view to ensure that the network can thoroughly learn the original feature information. Therefore, we introduce the reconstruction loss:
\begin{equation}
	\label{loss_rec}
	\mathcal{L}_{\text{Re}c}=\sum_{m=1}^M{\lVert \mathbf{X}^{(m)}-Dec_m( Enc_m( \mathbf{X}^{(m)}) ) \rVert _{2}^{2}},
\end{equation}
where $Enc_m$ and $Dec_m$ denote the encoder and decoder. The introduction of reconstruction loss makes the representation retain more view information while avoiding trivial solutions.

\subsection{Information Bottleneck}
Inspired by information bottleneck theory, we want to obtain the common semantics across all views and avoid the meaningless view-private information. For $\mathcal{L}_{IB}$, we have the following formula:
\begin{equation}
	\begin{aligned}
		\mathcal{L}_{IB}=\sum_{m=1}^M{( I(\mathbf{Z}^{(m)},\mathbf{Z}) -\beta I( \mathbf{Z}^{(m)},\mathbf{X}^{(m)}))},
	\end{aligned}
	\label{Lib}
\end{equation}
where $\mathbf{Z}^{(m)}$ denotes the within-view representation matrix of $m$-th view. For the consistent representation $\mathbf{Z}$, we first initialize it randomly and then let $\mathbf{Z}$ be updated through the network $f_{{\theta_{con}}}$. The first term of Eq.(\ref{Lib}) captures the common semantics of views by maximizing the mutual information of $\mathbf{Z}^{(m)}$ and $\mathbf{Z}$. The second term of Eq.(\ref{Lib}) erase the learned redundant information by minimizing the $\mathbf{Z}^{(m)}$ and ${\mathbf{X}}^{(m)}$. According to the computational formula of mutual information \cite{alemi2017deep}, $\mathcal{L}_{IB}$ can be written as the following formula:
\begin{equation}
	\begin{aligned}
		\mathcal{L}_{I B}=&\sum_{m=1}^M \sum_{\mathbf{z}^{(m)}} \sum_{\mathbf{z}} {p}(\mathbf{z}^{({m})}, \mathbf{z}) \log \frac{{p}(\mathbf{z} \mid \mathbf{z}^{({m})})}{{p}(\mathbf{z})} \\
		&-\beta \sum_{m=1}^M \sum_{\mathbf{z}^{({m})}} \sum_{\mathbf{x}^{({m})}} {p}(\mathbf{z}^{({m})}, \mathbf{x}^{({m})}) \log \frac{{p}(\mathbf{x}^{({m})} \mid \mathbf{z}^{({m})})}{{p}(\mathbf{x}^{({m})})},
	\end{aligned}
\end{equation}
since it is intractable in our case, we introduce the variational approximation ${q}( \mathbf{z}|\mathbf{z}^{(m)} )$ to ${p}( \mathbf{z}|\mathbf{z}^{(m)} )$. Because of the fact that Kullback Leibler divergence is always positive, we have:
\begin{equation}
	\begin{aligned}
		&\text{KL}({p}(\mathbf{z}|\mathbf{z}^{(m)}) ||{q}( \mathbf{z}|\mathbf{z}^{(m)})) =\sum_{\mathbf{z}}{p(\mathbf{z}|\mathbf{z}^{({m})}) \log \frac{{p}( \mathbf{z}|\mathbf{z}^{({m})} )}{q( \mathbf{z}|\mathbf{z}^{({m})})}}\ge 0 \\
		&\Rightarrow \sum_{\mathbf{z}}{p(\mathbf{z}|\mathbf{z}^{({m})})}\log {p}(\mathbf{z}|\mathbf{z}^{({m})})\ge \sum_{\mathbf{z}}{{p}(\mathbf{z}|\mathbf{z}^{({m})})}\log {q}( \mathbf{z}|\mathbf{z}^{({m})} ),
	\end{aligned}
\end{equation}
and hence the lower bound on the mutual information of the feature representation $\mathbf{z}^{( {m} )}$ and the consistent representation $\mathbf{z}$ can be computed as:
\begin{equation}
	\label{eq_z_iz}
	\begin{aligned}
		&I( \mathbf{Z}^{(m)},\mathbf{Z}) \ge \sum_{\mathbf{z}^{( \mathbf{m} )}}{\sum_{\mathbf{z}}{{p}( \mathbf{z}^{({m})},\mathbf{z}) \log \frac{{q}( \mathbf{z}|\mathbf{z}^{( {m} )} )}{{p}( \mathbf{z})}}} \\
		&=\sum_{\mathbf{z}^{({m})}}{\sum_{\mathbf{z}}{{p}( \mathbf{z}^{({m})},\mathbf{z} )}}\log {q}( \mathbf{z}|\mathbf{z}^{( {m} )} ) + H( \mathbf{z}).
	\end{aligned}
\end{equation}
Since the entropy of the $H(\mathbf{z})$ is independent of the optimization process, it can be ignored. Similar to Eq.(\ref{eq_z_iz}) , we consider the term $I(\mathbf{Z}^{(m)},\mathbf{X}^{(m)})$ as:
\begin{equation}
	\begin{aligned}
		&I( \mathbf{Z}^{(m)},\mathbf{X}^{(m)} ) \le \sum_{\mathbf{x}^{( {m} )}}{\sum_{\mathbf{z}^{( {m} )}}{\mathbf{p}( \mathbf{x}^{( {m} )},\mathbf{z}^{( {m} )} )}}\log \frac{{p}( \mathbf{z}^{( {m} )}|\mathbf{x}^{( {m} )} )}{{q}( \mathbf{z}^{( {m} )} )} \\
		&=\sum_{\mathbf{x}^{({m})}}{{p}(\mathbf{x}^{({m})})}\text{KL}({p}(\mathbf{z}^{({m} )}|\mathbf{x}^{({m})} ) ||{q}( \mathbf{z}^{({m})})).
	\end{aligned}
\end{equation}

\subsection{Multi-view Semantic Consistency}
As shown above, we compress the feature data by the information bottleneck theory, which ensures that the learned representation has a discriminative representation for the samples through the within-view reconstruction. In order to mine the semantic information of feature representations and enhance the model's ability to learn the consistency of different views, we introduce a semantic consistency learning module. Specifically, we use MLP network $f_{{\theta _{sem}}}$ to obtain the semantic information $\{ \mathbf{Q}^{{(m)}}\in \mathbb{R}^{N\times K} \} _{i=m}^{M}$ of the representation feature, $\mathbf{q}_{{ij}}^{{(m)}}$ denotes the $i$-th sample belongs to the $j$-th cluster in the $m$-th view. 

Naturally, we believe that the semantic embeddings of different views of the same sample should have a closer distance in the semantic space, and the semantic embeddings of different samples should be further apart. Inspired by \cite{chen2020simple}, we utilize the cosine similarity to measure the similarity between two semantic vectors:
\begin{equation}
	d ( \mathbf{q}_{\cdot {i}}^{( {m} )},\mathbf{q}_{\cdot {j}}^{( {n} )} ) =\frac{\mathbf{q}_{\cdot {i}}^{( {m} )}\cdot \mathbf{q}_{\cdot {j}}^{( {n} )}}{\lVert \mathbf{q}_{\cdot {i}}^{( {m} )} \rVert \lVert \mathbf{q}_{\cdot {j}}^{( {n} )} \rVert}.
\end{equation}
Then, we define the semantic consistency contrastive loss between $\mathbf{q}^{{(m)}}$ and $\mathbf{q}^{{(n)}}$ as:
\begin{equation}
	\begin{aligned}
		{l}_{sc}^{(mn)} =  - \frac{1}{K}\sum\limits_{j = 1}^K {\log } \frac{{{e^{d({\mathbf{q}_{ \cdot j}^{(m)},\mathbf{q}_{ \cdot j}^{(n)}})/{\tau}}}}}{{\sum\limits_{k = 1}^K {\sum\limits_{v = m,n} {( {{e^{d( {\mathbf{q}_{ \cdot j}^{(m)},\mathbf{q}_{ \cdot k}^{(v)}})/{\tau}}}}) - {e^{1/{\tau}}}} } }},
	\end{aligned}
	\label{eq9}
\end{equation}
where $\tau$ is the temperature parameter. In order to constrain the consistent representation $\mathbf{z}$ by semantic consistency, we add the semantic consistency contrastive loss term between $\mathbf{q}^{(m)}$ and $\mathbf{q}^{(c)}$:
\begin{equation}
	\begin{aligned}
 		{l}_{sc}^{(mc)}=-\frac{1}{K}\sum\limits_{j=1}^K{\log}\frac{e^{d( \mathbf{q}_{\cdot j}^{( m )},\mathbf{q}_{\cdot j}^{( c )} ) /\tau}}{\sum\limits_{k=1}^K{( e^{d( \mathbf{q}_{\cdot j}^{( c )},\mathbf{q}_{\cdot k}^{( m )} ) /\tau} ) -e^{1/\tau}}},
	\end{aligned}
	\label{eq10}
\end{equation}
where $\mathbf{q}_{{ij}}^{{(c)}}$ denotes the $i$-th sample belongs to the $j$-th cluster in the consistent representation $\mathbf{z}$. Because Eq.(\ref{eq9}) and Eq.(\ref{eq10}) compute the consistent contrastive loss in the same semantic space, we use the same parameters $\tau$ here. Thus, the full loss of semantic consistency $\mathcal{L}_{Sem}$ is defined as follows:
\begin{equation}
	\label{L_sem}
	\mathcal{L}_{Sem}=\frac{1}{2}\sum\limits_{m=1}^M{( \sum_{n\ne m}{{l}_{sc}^{( mn )}+}{l}_{sc}^{( mc )} )}+\mathcal{L}_{reg},
\end{equation}
where the firet term denotes the contrastive loss of the whole datasets. The second term $\mathcal{L}_{reg}$ indicates the regularization phase, which is usually done with an entropy loss to avoid all samples being assigned to a single cluster:
\begin{equation}
	{\mathcal{L}_{reg}} = \sum\limits_{m = 1}^M {\sum\limits_{j = 1}^K {\frac{1}{N}\sum\nolimits_{i = 1}^N {\mathbf{q}_{ij}^{(m)}} } } \log ( {\frac{1}{N}\sum\nolimits_{i = 1}^N {\mathbf{q}_{ij}^{(m)}} } ).
\end{equation}

\begin{algorithm}[tbp]
	\renewcommand{\algorithmicrequire}{\textbf{Input:}}
	\renewcommand{\algorithmicensure}{\textbf{Output:}}
	\caption{Optimization algorithm of MSCIB}
	\label{alg1}
	\begin{algorithmic}[1]
		\REQUIRE Multi-view dataset $\{ \mathbf{X}^{(m)} \} _{m=1}^{M}$; Parameter $\lambda_1$; $\lambda_2$; Trade-off factor $\beta$; The number of categories $K$.
		\STATE Initialize $\{ f_{en}^{\mu (m)},f_{en}^{\sigma (m)},f_{de}^{(m)}\} _{m=1}^{M}$ by minimizing Eq.(\ref{loss_rec}).
		\STATE Random Initialize $\mathbf{Z}$;
		\WHILE{not coverage}
		\STATE Obtain the within-view feature representation $\mathbf{z}^{(m)}$ by varitional autoencoder Eq.(\ref{p_z_x}) and reparameterization trick Eq.(\ref{reparameter});
		\STATE Obtain semantic label $\{ \mathbf{Q}^{(m)} \} _{m=1}^{M}$;
		\STATE Update network $\{ f_{en}^{\mu (m)},f_{en}^{\sigma (m)},f_{de}^{(m)},f_{sem}^{(m)} \} _{m=1}^{M}$, $f_{{\theta_{con}}}$ and representation matrix $\mathbf{Z}$ based on Eq.(\ref{loss_all}), (\ref{loss_rec}), (\ref{L_sem}), (\ref{loss_IB}).
		\ENDWHILE
		\ENSURE  The Consistent Feature Representation $\mathbf{Z}$.
	\end{algorithmic}
        \label{algorithm}
\end{algorithm}

\subsection{Implementation Details}

Inspired by \cite{alemi2017deep}, we use a variational autoencoder $
{\mu _{\mathbf{z}|{\mathbf{x}}}} = f_{{\theta _{en}}}^\mu ( \mathbf{x} )$ and ${\sigma _{\mathbf{z}|{\mathbf{x}}}} = f_{{\theta _{en}}}^\sigma ( \mathbf{x} )$ to obtain the mean and variance of the representation features. For $p( {\mathbf{z}^{(m)}|{\mathbf{x}^{(m)}}} )$, we assume that:
\begin{equation}
	\label{p_z_x}
	p( {\mathbf{z}^{(m)}|{\mathbf{x}^{(m)}}} ) = \mathcal{N}( {{\mu _{\mathbf{z}^{(m)}|{\mathbf{x}^{(m)}}}},\sigma _{\mathbf{z}^{(m)}|{\mathbf{x}}^{(m)}}^2I} ).
\end{equation}
In order to learn the representation feature ${\mathbf{z}^{(m)}}$, we use the reparameterization trick \cite{Kingma2014}.
\begin{equation}
	\label{reparameter}
	z = {\mu _{\mathbf{z}^{(m)}|{\mathbf{x}^{(m)}}}} + {\sigma _{\mathbf{z}^{(m)}|{\mathbf{x}}^{(m)}}} \odot \varepsilon,
\end{equation}
where $\boldsymbol{\varepsilon }\sim \mathcal{N}( 0,I )$, and $\odot$ is the element-wise product. The MLP network $f_{\theta _{de}}(\mathbf{z})$ is used to construct the decoder. For the information bottleneck term, we rewrite $p( \mathbf{z},\mathbf{z}^{(m)}) dz=p( \gamma ) d\gamma $, where $\mathbf{z}=g_1( z,\gamma )$ with the gaussian random variable $\gamma$. Therefore, we can rewrite the information bottleneck loss term as:
\begin{equation}
	\label{loss_IB}
	\begin{aligned}
		{\mathcal{L}_{IB}} &= \frac{1}{N}\sum_{i = 1}^N \sum_{m = 1}^M \{  -\mathbb{E}_{\gamma }\log q({\mathbf{z}}|g({\mathbf{z}^{(m)}},\gamma )) \\
		&+ \beta KL(p({\mathbf{z}^{(m)}}|{x^{(m)}})||q({\mathbf{z}^{(m)}}))\},
	\end{aligned}
\end{equation}
where the first term is reconstruction loss, and the second term is the KL divergence. In this way, irrelevant redundant information in the original view can be reduced and a compact and comprehensive multi-view information bottleneck representation can be obtained. The complete optimization process of MSCIB is summarized in Algorithm \ref{algorithm}, specifically, our model consists of variational autoencoder, feature decoder, and semantic MLP.

\subsection{Comparison with Existing MVC Methods based on IB}
The existing methods of applying information bottleneck to multi-view data are mainly expanded by the following:
\begin{equation}
	\label{IB}
	\begin{aligned}
		\mathop {\max }\limits_{Z,{Z^{(1)}},{Z^{(2)}},...{Z^{(M)}}} I(Y,Z) - \sum\limits_{m=1}^M {{\alpha ^{(m)}}I({X^{(m)}},{Z^{(m)}})},
	\end{aligned}
\end{equation}
where $\alpha$ is the balance parameter. Since multi-view clustering is an unsupervised task, the existing multi-view clustering methods based on information bottleneck mainly focus on the construction of $\mathbf{Y}$. CMIB\cite{wan2021multi} learns the common information between views by maximizing the mutual information of the multi-view information bottleneck representation $\mathbf{Z}$ and the common representation $\mathbf{H}$. However, due to the complexity of multi-view data, the feature representation is not very discriminative, and mutual information loss is difficult to reduce adverse information.

Compared with previous methods, the main difference is that MSCIB maps feature representations to semantic space, where the semantic consistency operation can decrease the distance between the same sample in multi-view and increase the distance between different samples. In this way, the obtained sample representations have strong discriminability. Under the guidance of information bottleneck theory, MSCIB is more likely to learn the sample representation with consistent information.

\section{Experiments}
\subsection{Experiment Setup}
\subsubsection{Datasets}
Our experiments are conducted on eight datasets, the details of which are shown in Table \ref{dataset}. \textbf{MNIST-USPS}\cite{pmlr-v97-peng19a} is a popular dataset of handwritten digits that contains 5,000 samples of digit images with two different styles. \textbf{BDGP}\cite{bioinformatics} contains 2500 drosophila embryo samples, each with visual and textual features. \textbf{Caltech}\cite{FEIFEI200759} collects 1400 images belonging to 7 categories with 5 views. Here four sub-datasets, Caltech-2V, Caltech-3V, Caltech-4V and Caltech-5V with different number of views are constructed for evaluation. Specifically, Caltech-2V uses WM and centrist; Caltech-3V uses WM, CENTRIST and Lycium barbarum polysaccharide; Caltech-4V uses WM, CENTRIST, LBP, and GIST; Caltech-5V uses WM, CENTRIST, LBP, GIST, and HOG. In \textbf{Multi-MNIST}, samples with different views mean the same number written by different people. \textbf{Noisy-MNIST}\cite{COMPLETER} contains 50000 samples of digit images, which uses the original MNIST as view 1 and randomly selects within-class images with white Gaussian noise as view 2.

\subsubsection{Comparing Methods}
We compare MSCIB against the following popular and state-of-the-art methods. \textbf{MFLVC}\cite{MFLVC} proposes a new framework of multi-level feature learning for contrastive multi-view clustering to address the clustering issue. $\text{\textbf{AE}}^{2}\text{\textbf{-NET}}$\cite{AE2NET} integrates information from heterogeneous sources into a complete representation via a nested autoencoder framework. \textbf{DEMVC}\cite{DEMVCINS2021} proposes a novel co-training multi-view clustering scheme. \textbf{SDMVC}\cite{SDMVC} uses a deep autoencoder to learn the embedded features for each view independently. \textbf{GUMRL}\cite{GUMRL} performs view-specific feature representation learning guided by graph information. \textbf{CUMRL}\cite{CUMRLTCSVT2021} introduces a collaborative learning strategy to link view-oriented compact learning with unified representation learning in CUMRL by exploiting the low-rank constraint of tensors. As an unsupervised representation learning algorithm, \textbf{CMIB}\cite{wan2021multi} relies on the information bottleneck theory, integrating shared representations between different views and view-specific representations for each view.

\begin{table}[tbp]
	\centering
	\begin{tabular}{c|c|c|c}
		\toprule
		Datasets & \#Samples & \#Views & \#Classes \\
		\midrule
		MNIST-USPS & 5000  & 2     & 10 \\
		BDGP  & 2500  & 2     & 5 \\
		Multi-MNIST & 70000 & 2     & 10 \\
		NoisyMNIST & 50000 & 2     & 10 \\
		Caltech-2V & 1400  & 2     & 7 \\
		Caltech-3V & 1400  & 3     & 7 \\
		Caltech-4V & 1400  & 4     & 7 \\
		Caltech-5V & 1400  & 5     & 7 \\
		\bottomrule
	\end{tabular}%
	\caption{The information of the datasets in our experiments.}
	\label{dataset}%
\end{table}%

\subsubsection{Implementation}
All datasets are constructed as one-dimensional vectors for manipulation, and fully connected networks are employed to implement the autoencoder for all views in our MSCIB. For all datasets, the ReLU activation function is used to implement variational autoencoders in MSCIB. Adam optimizer \cite{kingma2014adam} is used for optimization. Our method is implemented by PyTorch on one NVIDIA Geforce GTX 1080ti GPU with 11GB memory. For the comparison methods, we utilize the code published by its corresponding author to run on our machine and use the recommended settings in its original work.

\subsubsection{Evaluation metrics}
We use $k$-means to perform the clustering task on the feature consistent representation. Four metrics evaluate the clustering effectiveness, i.e., clustering accuracy (ACC), normalized mutual information (NMI) and adjusted rand index (ARI). For each dataset, the mean values of 10 runs are reported for all methods.
\begin{table*}[tbp]
	\centering
	\resizebox{1.0\textwidth}{!}{
		\begin{tabular}{r|ccc|ccc|ccc|ccc}
			\toprule
			Datasets & \multicolumn{3}{c}{MNIST-USPS} & \multicolumn{3}{c}{BDGP} & \multicolumn{3}{c}{Multi-MNIST} & \multicolumn{3}{c}{Noisy-MNIST} \\
			\midrule
			Evaluation & ACC   & NMI   & ARI   & ACC   & NMI   & ARI   & ACC   & NMI   & ARI   & ACC   & NMI   & ARI \\
			\midrule
			$\text{AE}^2$NET(2019) & 0.7738 & 0.7487 & 0.9385 & 0.6350 & 0.5097 & 0.6781 & 0.3068 & 0.2311 & 0.1593 & 0.1690 & 0.0611 & 0.0259 \\
			DEMVC(2021) & 0.7786 & 0.9014 & 0.7774 & 0.5736 & 0.4727 & 0.4137 & 0.9996 & 0.9983 & 0.9998 & 0.2557 & 0.2331 & 0.0930 \\
			CUMRL(2021) & 0.7698 & 0.7755 & 0.9417 & 0.7305 & 0.5906 & 0.8085 & 0.7038 & 0.7398 & 0.9302 & 0.2691 & 0.2418 & 0.8232 \\
			SDMVC(2021) & 0.9368 & 0.9431 & 0.9825 & 0.9652 & 0.9086 & 0.9731 & 0.9988 & 0.9957 & 0.9995 & 0.2633 & 0.2462 & 0.0823 \\
			CMIB(2021) & 0.6399 & 0.6346 & 0.5057 & 0.2427 & 0.0129 & 0.0079 & 0.8396 & 0.8062 & 0.7633 & 0.4258 & 0.4135 & 0.2859 \\
			GUMRL(2022) & 0.8396 & 0.8379 & 0.9602 & 0.8213 & 0.6719 & 0.8555 & 0.7941 & 0.7686 & 0.7119 & 0.5659 & 0.5141 & 0.4070 \\
			MFLVC(2022) & 0.9956 & 0.9865 & 0.9902 & 0.9872 & 0.9613 & 0.9684 & 0.9994 & 0.9984 & 0.9991 & 0.9892 & 0.9686 & 0.9766 \\
			\textbf{MSCIB(Our)} & \textbf{0.9964} & \textbf{0.9893} & \textbf{0.9920} & \textbf{0.9924} & \textbf{0.9746} & \textbf{0.9811} & \textbf{0.9996} & \textbf{0.9985} & \textbf{0.9992} & \textbf{0.9957} & \textbf{0.9853} & \textbf{0.9905} \\
			\bottomrule
	\end{tabular}}
	\caption{Clustering results of all methods on MNIST-USPS, BDGP, Multi-MNIST, Noisy-MNIST. Bold denotes the best results.}
	\label{tab1}%
\end{table*}%

\begin{table*}[tbp]
	\centering
	\resizebox{1.0\textwidth}{!}{
		\begin{tabular}{r|ccc|ccc|ccc|ccc}
			\toprule
			Datasets & \multicolumn{3}{c}{Caltech-2V} & \multicolumn{3}{c}{Caltech-3V} & \multicolumn{3}{c}{Caltech-4V} & \multicolumn{3}{c}{Caltech-5V} \\
			\midrule
			Evaluation & ACC   & NMI   & ARI   & ACC   & NMI   & ARI   & ACC   & NMI   & ARI   & ACC   & NMI   & ARI \\
			\midrule
			$\text{AE}^2$NET(2019) & 0.4614 & 0.3201 & 0.2037 & 0.5148 & 0.4108 & 0.3124 & 0.4801 & 0.3889 & 0.2404 & 0.6767 & 0.5813 & 0.3524 \\
			DEMVC(2021) & 0.4050 & 0.2573 & 0.1337 & 0.4721 & 0.2914 & 0.2254 & 0.4500  & 0.2795 & 0.2101 & 0.5264 & 0.4004 & 0.3134 \\
			CUMRL(2021) & 0.5511 & 0.4469 & 0.3584 & 0.5295 & 0.4687 & 0.3678 & 0.7162 & 0.6655 & 0.6012 & 0.7286 & 0.6926 & 0.6279 \\
			SDMVC(2021) & 0.4786 & 0.3742 & 0.3573 & 0.4164 & 0.3065 & 0.2106 & 0.4350 & 0.3052 & 0.2477 & 0.4214 & 0.2824 & 0.2099 \\
			CMIB(2021) & 0.4208 & 0.2994 & 0.1967 & 0.4329 & 0.3247 & 0.2142 & 0.2941 & 0.1158 & 0.0658 & 0.3020 & 0.1494 & 0.0913 \\
			GUMRL(2022) & 0.5142 & 0.5083 & 0.3826 & 0.7003 & \textbf{0.6543} & 0.5518 & 0.7100  & 0.6676 & 0.5644 & 0.6863 & 0.6563 & 0.5394 \\
			MFLVC(2022) & 0.5993 & 0.5155 & 0.4074 & 0.6350 & 0.5727 & 0.4686 & 0.6807 & 0.6296 & 0.5254 & 0.7514 & 0.6801 & 0.5997 \\
			\textbf{MSCIB(Our)} & \textbf{0.6350} & \textbf{0.5391} & \textbf{0.4460} & \textbf{0.7214} & 0.6168 & \textbf{0.5523} & \textbf{0.8029} & \textbf{0.6833} & \textbf{0.6313} & \textbf{0.8314} & \textbf{0.7293} & \textbf{0.6861} \\
			\bottomrule
	\end{tabular}}
	\caption{Clustering results of the learned consistent multi-view representation on $\text{Caltech-}n\text{V}$, where $n$ is selected from $\text{{2, 3, 4, 5}}$.}
	\label{tab2}%
\end{table*}%

\subsection{Clustering Results}
The representation matrices learned on the six datasets and the clustering results after $k$-means\cite{MacQueen1967SomeMF} are shown in Table \ref{tab1} and \ref{tab2}. We can observe that MSCIB achieves the best performance on most datasets compared to the other algorithms. In particular, on the Multi-MNIST large-scale clustering dataset and the noise dataset NoisyMNIST, with the introduction of the information bottleneck theory, the noisy information that affects the clustering effect is reduced, MSCIB has achieved good results. Our method uses the information bottleneck theory to discard the redundant information irrelevant to the task and learns more discriminative information. The introduction of the idea of semantic consistency improves the learning of view consistent information in information bottleneck theory. Since our method is to learn a more integrated representation matrix, more view data has a stronger guiding significance for clustering, and MSCIB is also suitable for multi-view tasks.

To verify our speculation, multiple views are instructive for learning the consistent representation matrix. From Table \ref{tab2}, As the number of views increases, our $k$-means clustering accuracy improves by 14\%, 11\% and 3\% respectively. The data of each different view of the Caltech dataset not only contains the public semantic information of samples but also has a large amount of view private information and noise information. This adverse information increases with the number of views, so our model discards much redundant information by minimizing the mutual information between the representation matrix and the samples feature. The purpose of simultaneously maximizing the mutual information between the consistent representation and the within-view representation is to enable the consistent representation to learn more view information. By fusing the semantic information from different views, our consistent representation learns more common semantics, and our method also obtains a large improvement over the sub-optimal method. These observations further validate the effectiveness of the proposed method.

\subsection{Visualization results}
To visualize the effectiveness of the learned consistent representation, we visualize the $\mathbf{Z}$ implemented in epochs 5, 10, 15, and 50 of the MSCIB learning process based on $t$-SNE \cite{van2008visualizing}. Taking the MNIST-USPS dataset as an example, the visualization results are shown in Figure \ref{t-sne}, where the first row represents the result of the consistent representation $\mathbf{Z}$ and the second row represents the result of $\mathbf{Z}^{(1)}$. Since our consistent representation within-view representation $\mathbf{Z}$ is randomly initialized, the visualization results of $t$-SNE are somewhat messy in the initial stage of training because $\mathbf{Z}$ does not fully learn the common semantic information of the views. However, $\mathbf{Z}^{(1)}$ is obtained from the within-view feature representation, which contains some common semantic information, and the visualization results of $t$-SNE are more clustered than $\mathbf{Z}$. It can be seen that with the iterative update of parameters, compared with the within-view representation, the clustering structure of the consistent representation matrix becomes clearer. As a consequence, the proposed method can obtain a discriminative feature representation with a good structure.

\subsection{Model Analysis}

\begin{figure}[htbp]
	\centering
	\includegraphics[scale=0.35]{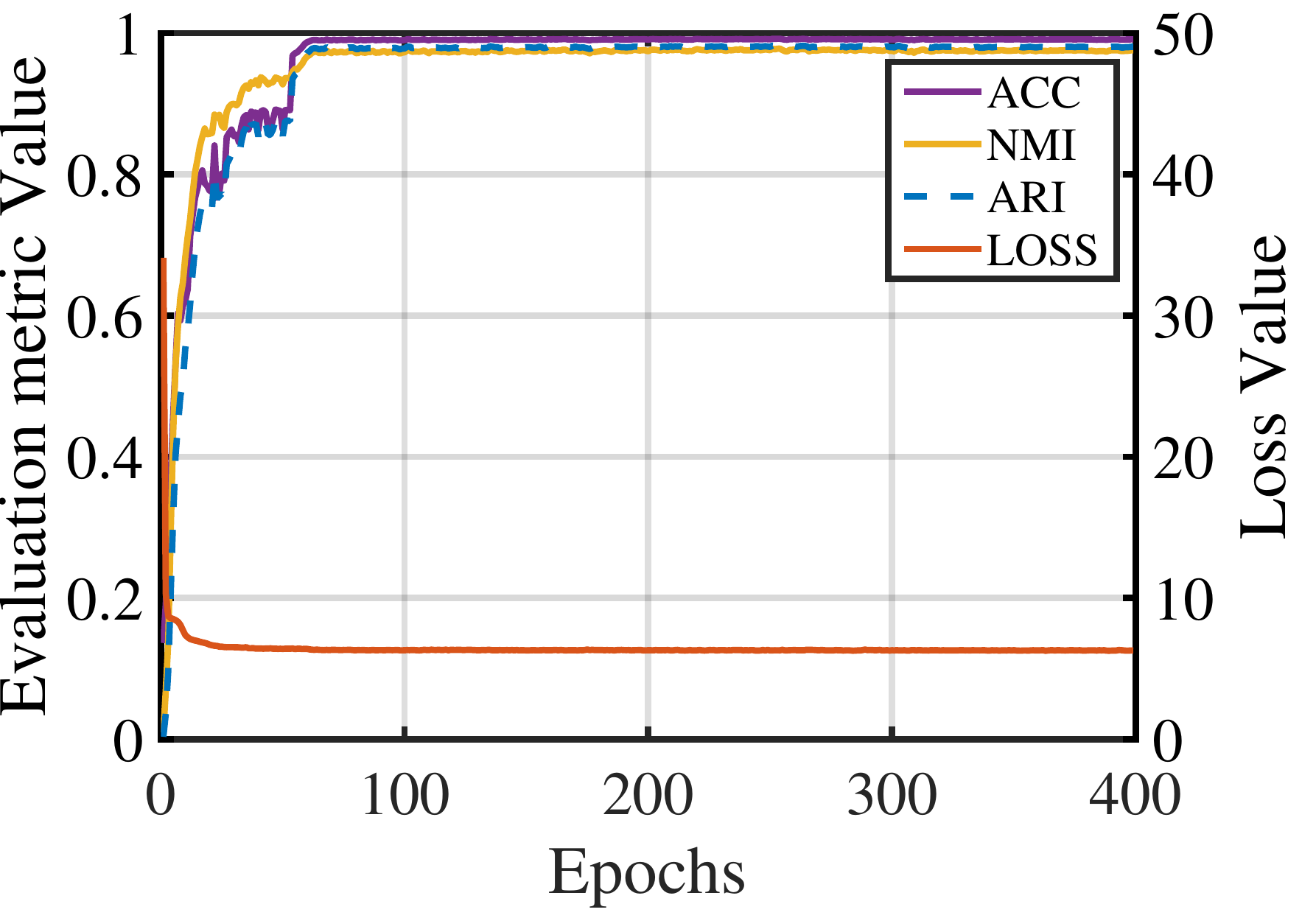}
	\caption{The visualization of the training process on MNIST-USPS.}
	\label{figloss}
\end{figure}

\subsubsection{Convergence analysis}
\begin{figure*}[ht]
	\centering
	\begin{subfigure}[t]{0.23\linewidth}
		\centering
		\includegraphics[width=0.8\linewidth, height=0.8\linewidth]{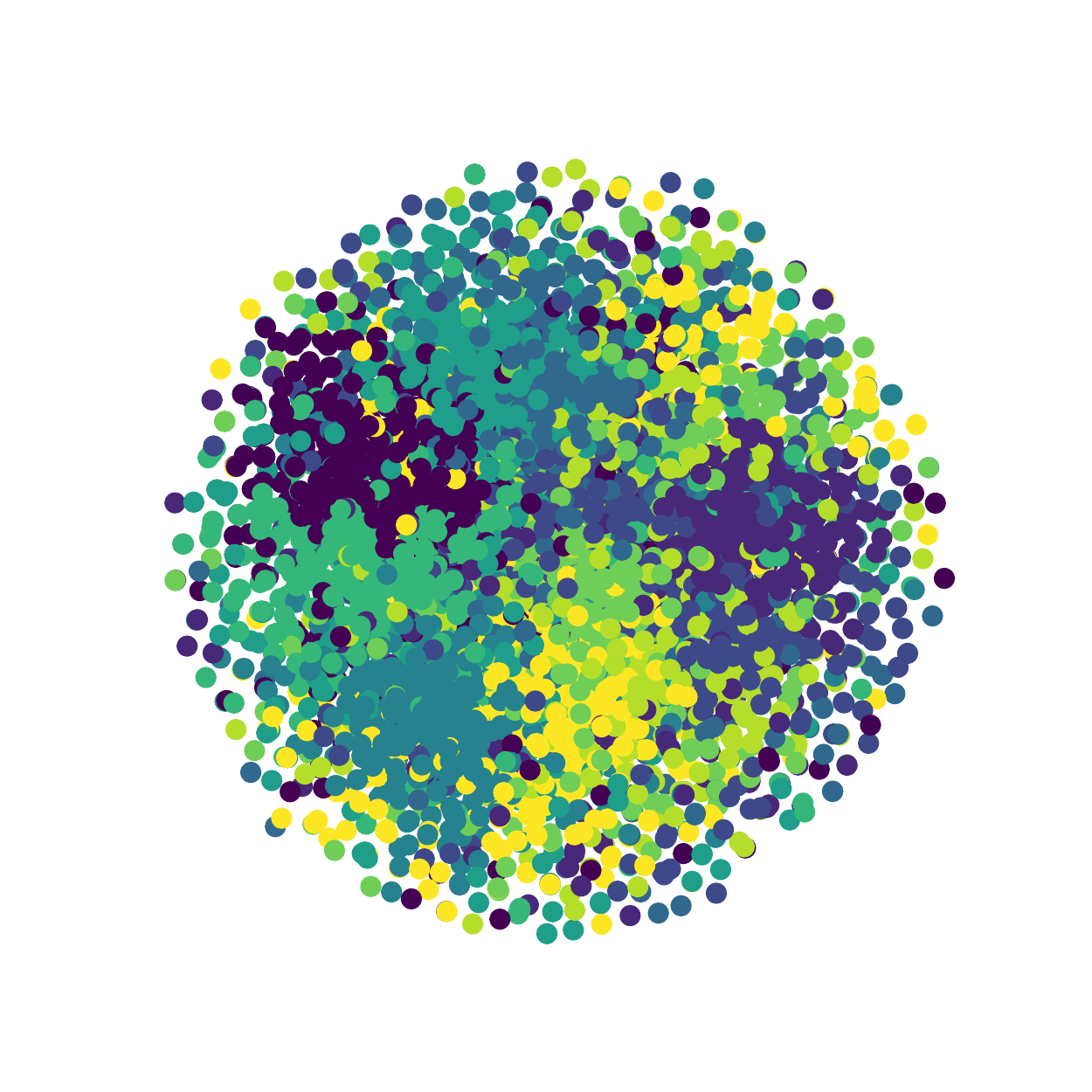}
		\caption{Epoch 5}
	\end{subfigure}
	\begin{subfigure}[t]{0.23\linewidth}
		\centering
		\includegraphics[width=0.8\linewidth, height=0.8\linewidth]{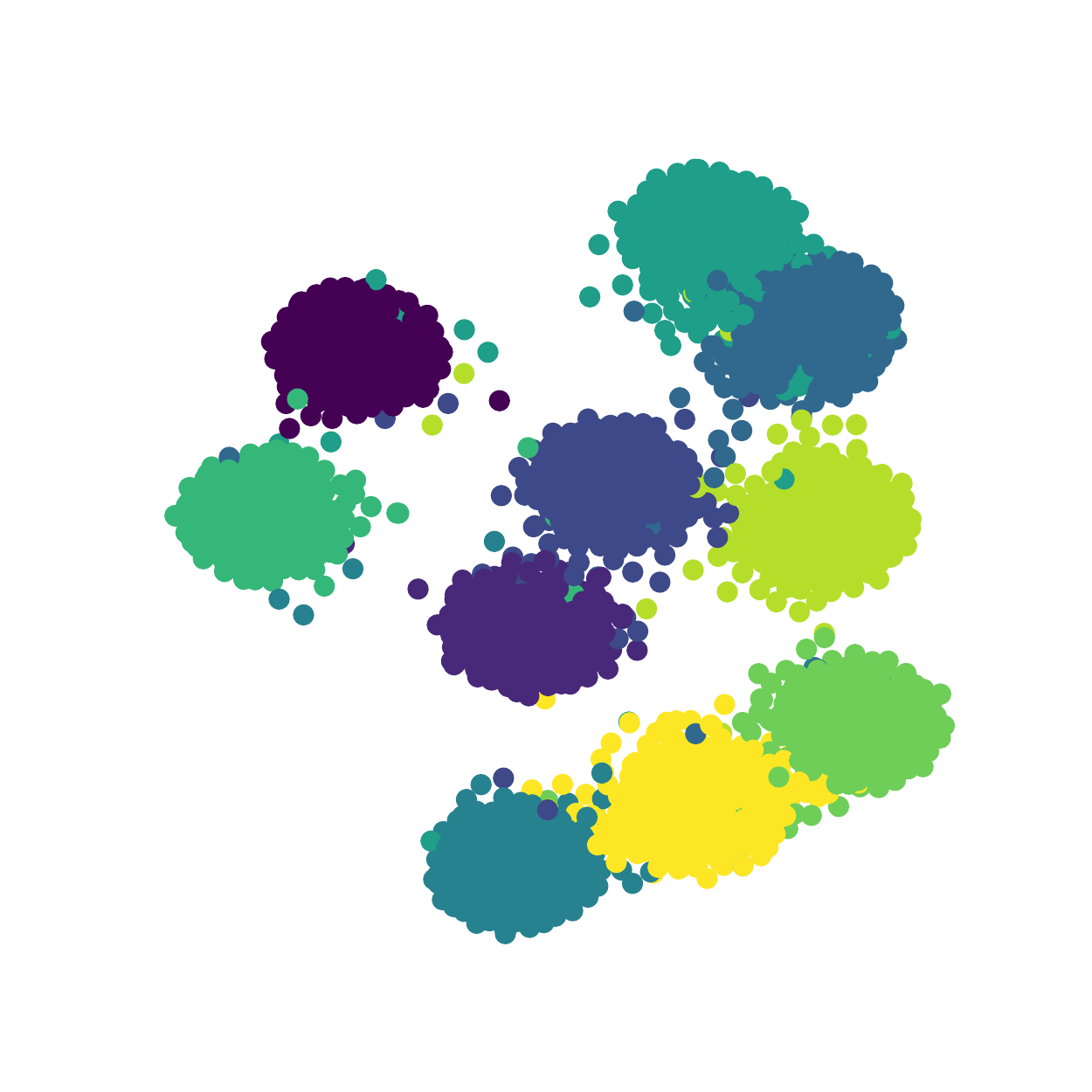}
		\caption{Epoch 10}
	\end{subfigure}
	\begin{subfigure}[t]{0.23\linewidth}
		\centering
		\includegraphics[width=0.8\linewidth, height=0.8\linewidth]{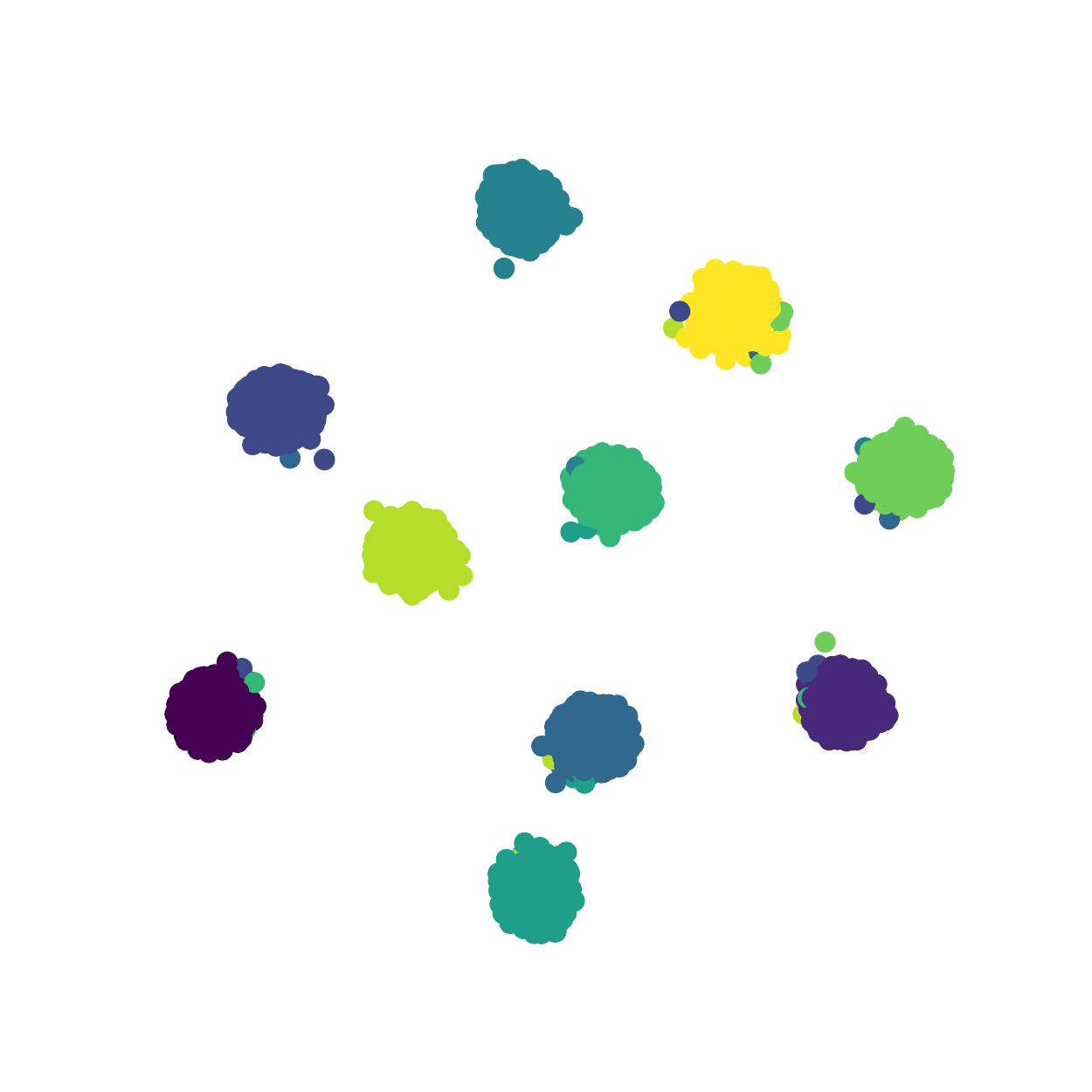}
		\caption{Epoch 15}
	\end{subfigure}
	\begin{subfigure}[t]{0.23\linewidth}
		\centering
		\includegraphics[width=0.8\linewidth, height=0.8\linewidth]{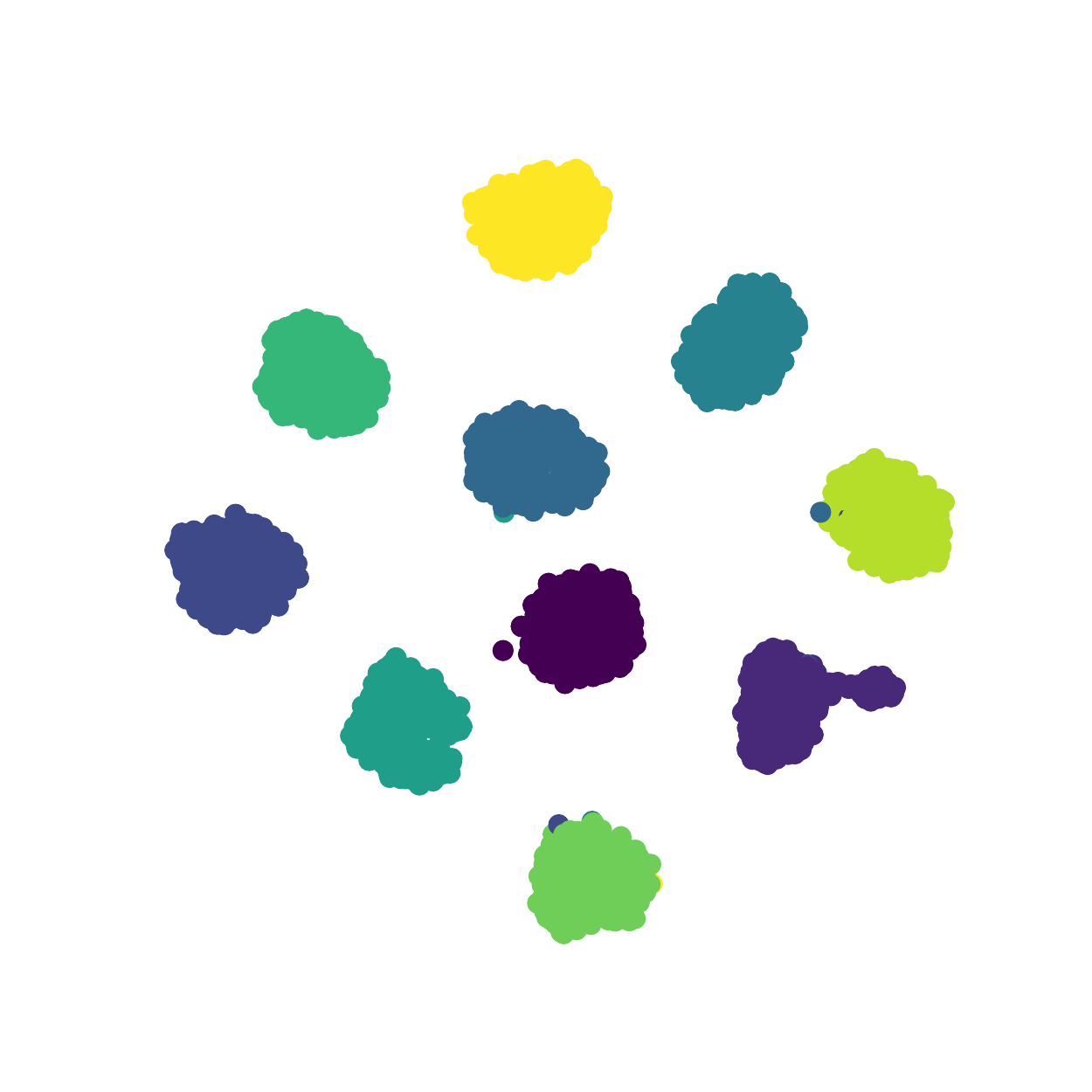}
		\caption{Epoch 50}
	\end{subfigure}\\
	\begin{subfigure}[t]{0.23\linewidth}
		\centering
		\includegraphics[width=0.8\linewidth, height=0.8\linewidth]{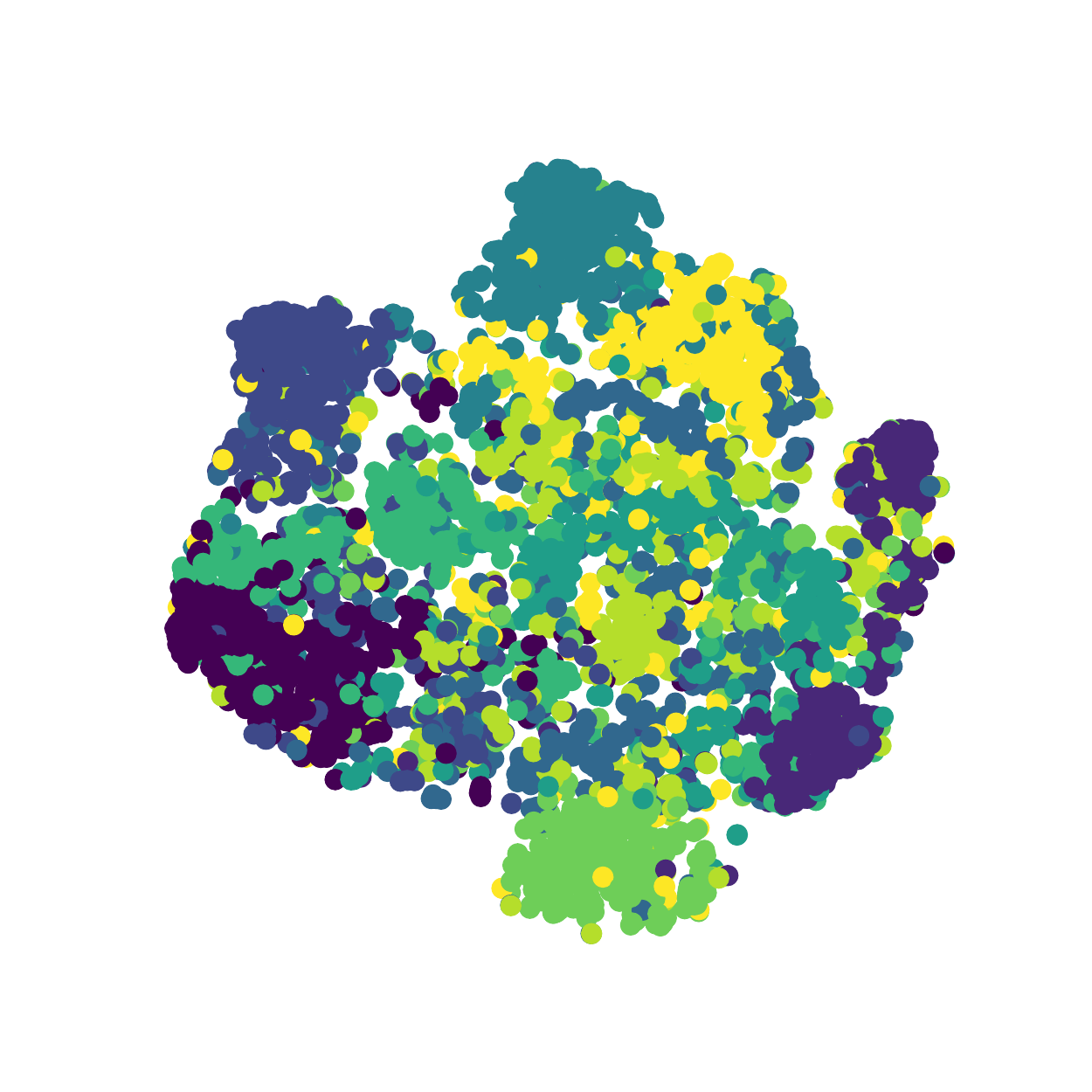}
		\caption{Epoch 5}
	\end{subfigure}
	\begin{subfigure}[t]{0.23\linewidth}
		\centering
		\includegraphics[width=0.8\linewidth, height=0.8\linewidth]{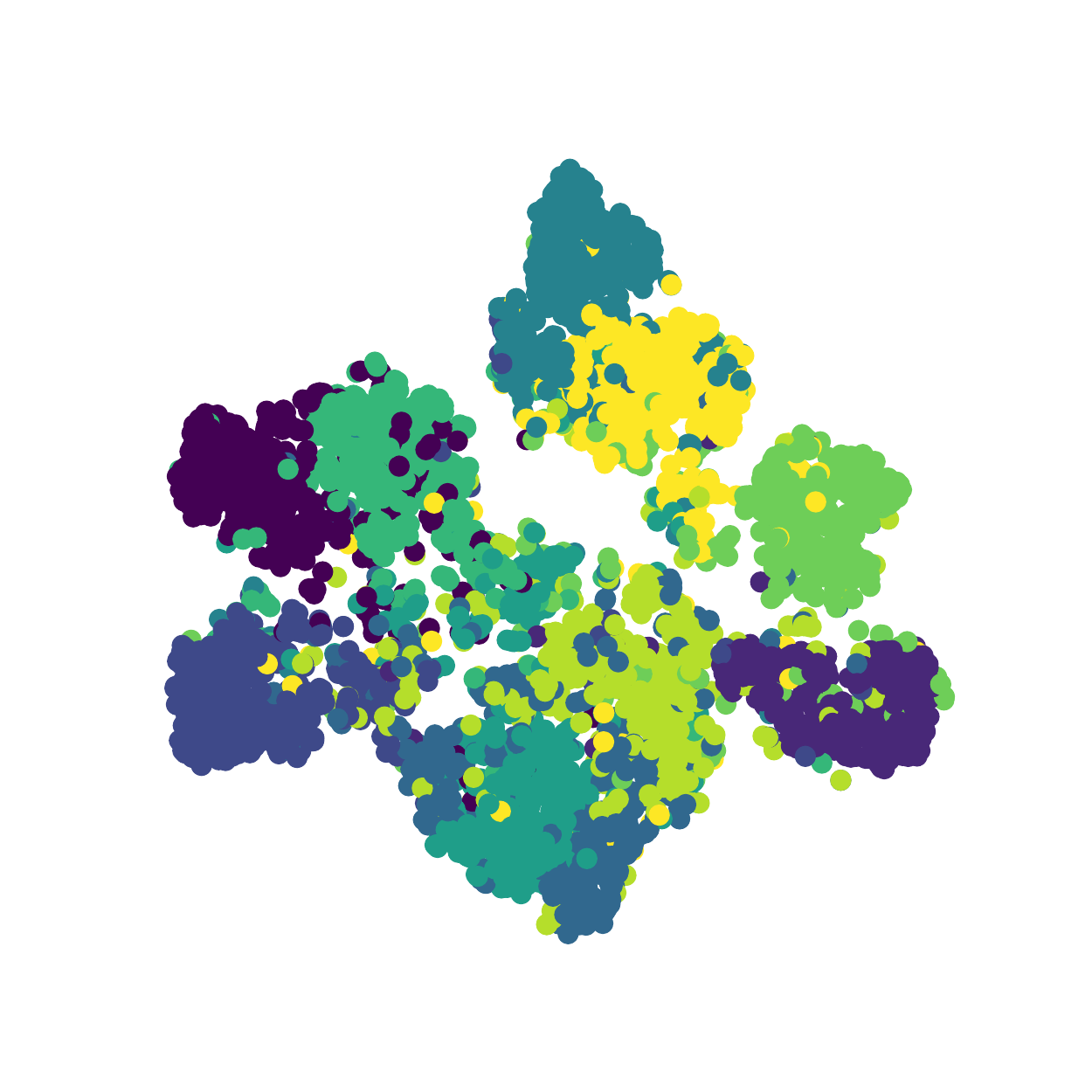}
		\caption{Epoch 10}
	\end{subfigure}
	\begin{subfigure}[t]{0.23\linewidth}
		\centering
		\includegraphics[width=0.8\linewidth, height=0.8\linewidth]{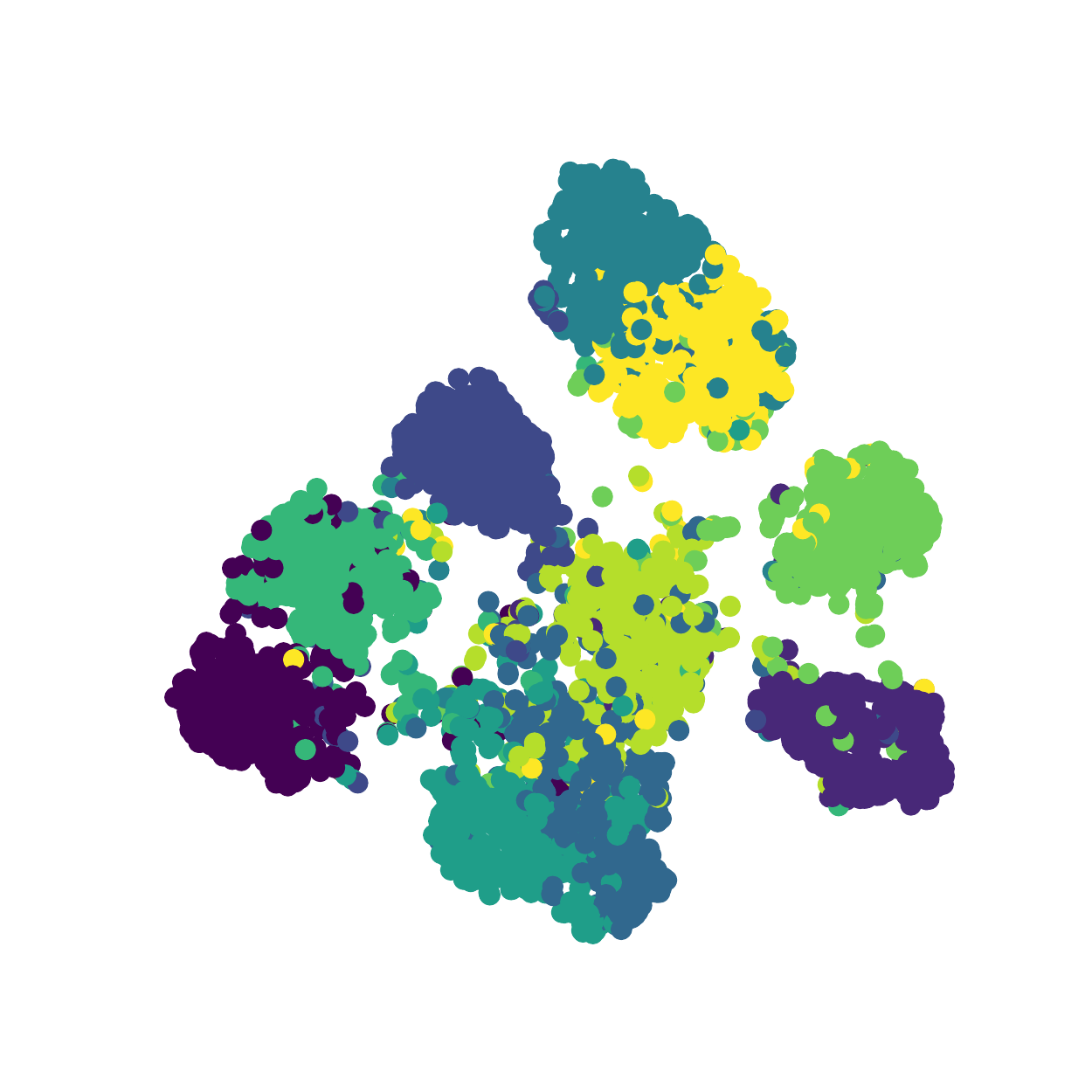}
		\caption{Epoch 15}
	\end{subfigure}
	\begin{subfigure}[t]{0.23\linewidth}
		\centering
		\includegraphics[width=0.8\linewidth, height=0.8\linewidth]{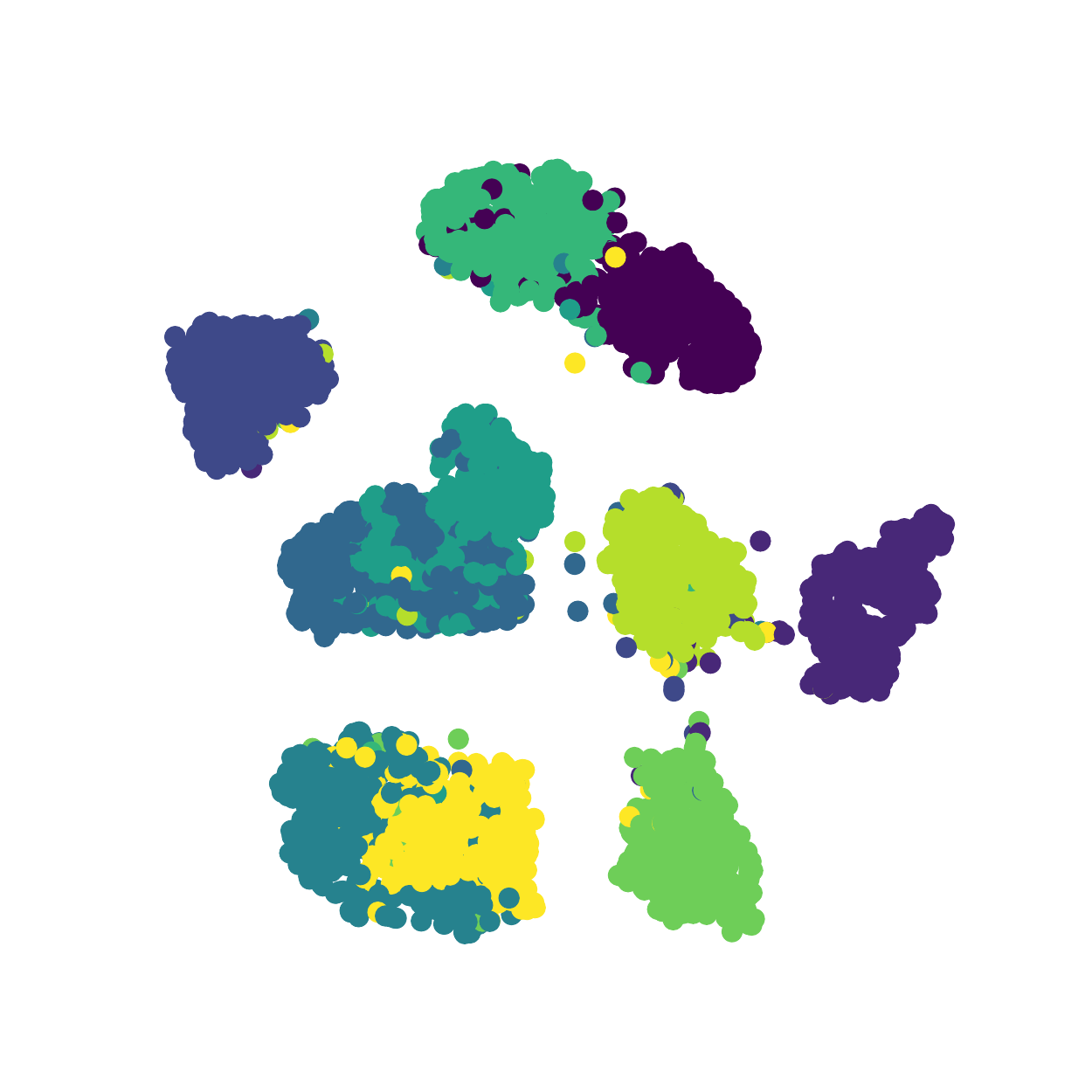}
		\caption{Epoch 50}
	\end{subfigure}\\
	\caption{The t-SNE visualization results of the consistent representation (a-d) and within-view representation (e-h) learned in Epoch 5, 10, 15, and 50.}
	\label{t-sne}
\end{figure*}
To verify the convergence of the proposed MSCIB model, we visualize the training process of MNIST-USPS as Figure \ref{figloss}. It can be observed that in the early stage of training, our LOSS drops sharply, ACC, NMI and ARI metrics rise sharply, and finally converge to good performance. For the other datasets, the convergence properties remain the same.

\subsubsection{Parameter sensitivity analysis}
\begin{figure}[ht]
	\centering
	\begin{subfigure}[t]{0.45\linewidth}
		\centering
		\includegraphics[width=1\linewidth, height=0.8\linewidth]{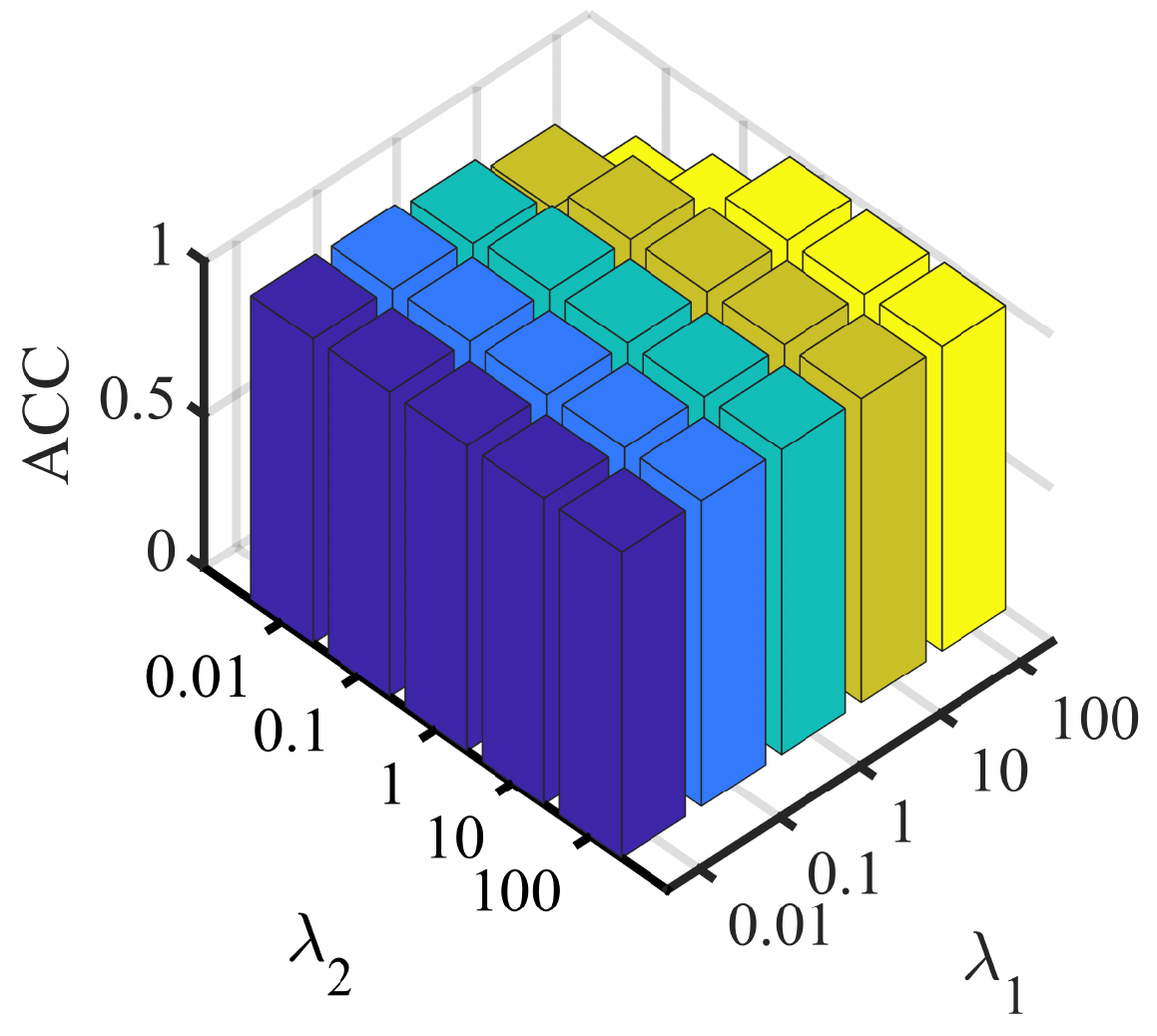}
		\caption{}
		\label{figbar1}
	\end{subfigure}
	\begin{subfigure}[t]{0.45\linewidth}
		\centering
		\includegraphics[width=1\linewidth, height=0.8\linewidth]{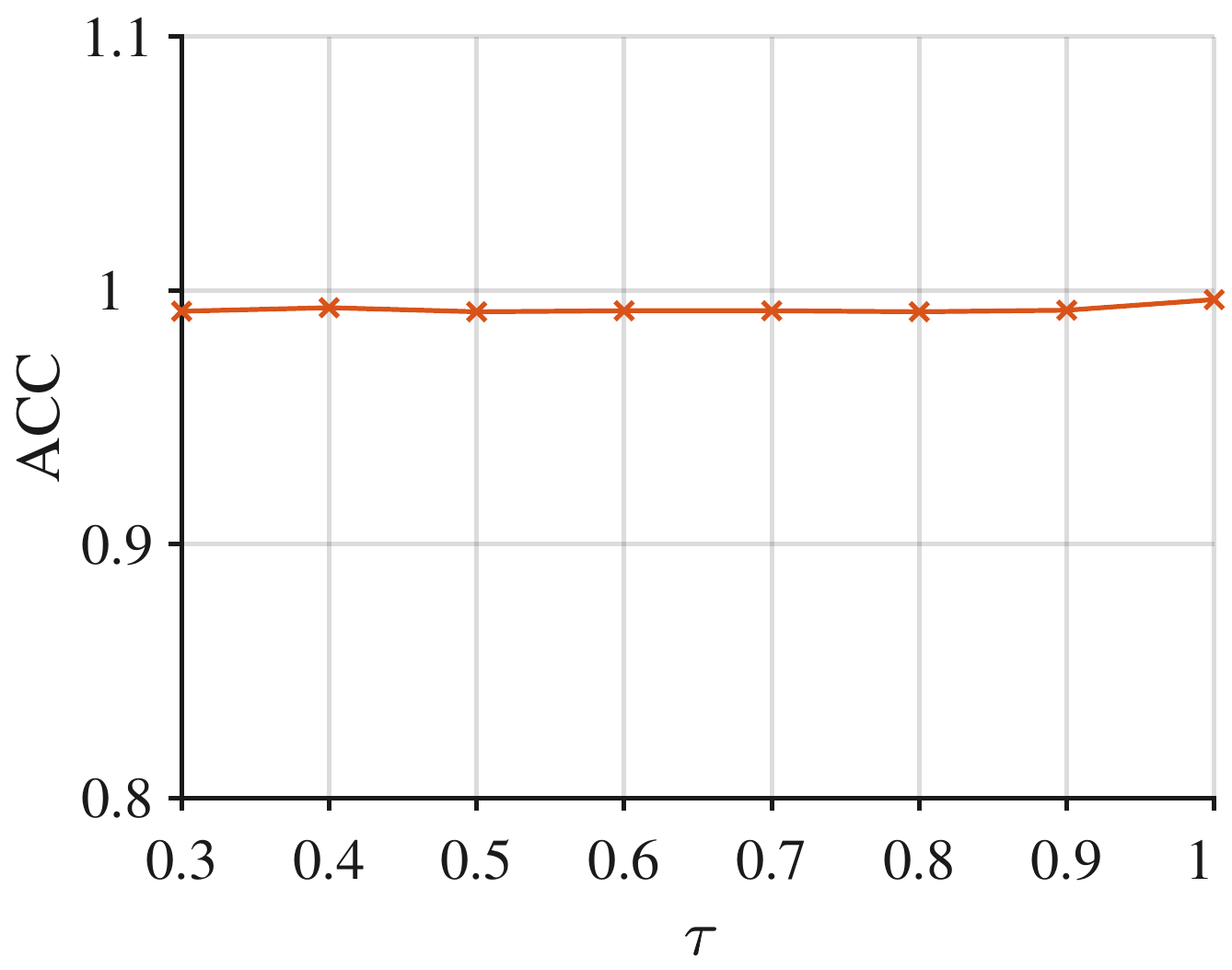}
		\caption{}
		\label{figbar2}
	\end{subfigure}
	\caption{Parameter sensitivity analysis on MNIST-USPS.}
	\label{figbar}
\end{figure}

To explore the sensitivity of MSCIB to hyperparameters, we take different values of $\lambda_1$ and $\lambda_2$ in the equation and explore their influence on clustering. To make the results more reliable, we ran the clustering task 10 times for averaging. The results are shown in Figure \ref{figbar1}, which illustrates that our model is not sensitive to $\lambda_1$ and $\lambda_2$. Therefore, $\lambda_1$ and $\lambda_2$ are set to 1 for all datasets in the experiments in the rest of the paper. The semantic consistency part also includes a temperature coefficient $\tau$. Figure \ref{figbar2} indicates that MSCIB is insensitive to the choice of $\tau$. Our method can achieve very good results, $\tau$ is set to 1 here.

\subsubsection{Ablation Studies}
To verify the effectiveness of introducing the information bottleneck theory and semantic consistency in our method, we do $k$-means clustering by using the original feature $ \mathbf{X}^{\left( m \right)}$, the view representation $\mathbf{Z}^{\left( m \right)}$, and the consistent representation $\mathbf{Z}$ respectively. The results of ablation experiments are shown in Table \ref{tab-ablation}, from which we can conclude that the learned multi-view consistent representation $\mathbf{Z}$ gives the best results.
\begin{table}[tbp]
	\centering
	\resizebox{0.48\textwidth}{!}{
		\begin{tabular}{c|c|c|c|c|c|c}
			\toprule
			Datasets & Evaluation & $\mathbf{X}^{(1)}$ & $\mathbf{X}^{(2)}$ & $\mathbf{Z}^{(1)}$ & $\mathbf{Z}^{(2)}$ & $\mathbf{Z}$ \\
			\midrule
			\multirow{2}[0]{*}{BDGP} & ACC   & 0.6376 & 0.6744 & 0.9600  & 0.9624 & \textbf{0.9652} \\
			& NMI   & 0.5149 & 0.5518 & 0.8893 & 0.9040 & \textbf{0.9086} \\
			\midrule
			\multirow{2}[0]{*}{\makecell[c]{MNIST\\ -USPS}} & ACC   & 0.5440 & 0.4787 & 0.9796 & 0.9916 & \textbf{0.9972} \\
			& NMI   & 0.4846 & 0.4690 & 0.9520 & 0.9774 & \textbf{0.9918} \\
			\midrule
			\multirow{2}[0]{*}{\makecell[c]{Multi\\ -MNIST}} & ACC   & 0.5401 & 0.5421 & 0.9870 & 0.9861 & \textbf{0.9988} \\
			& NMI   & 0.4816 & 0.4809 & 0.9687 & 0.9661 & \textbf{0.9957} \\
			\bottomrule
		\end{tabular}}%
	\caption{Ablation study of the original views and different representations for clustering task.}
	\label{tab-ablation}
\end{table}%

\section{Conclusion}
In this paper, we propose a novel multi-view clustering method that relies on the information bottleneck theory to reduce task-irrelevant redundant information and learn a more discriminative consistent representation. Meanwhile, the introduction of semantic consistency can well balance the consistency and complementarity between multiple views, so as to obtain a more reliable representation. Experiments are conducted on several real datasets, and experimental results illustrate that MSCIB achieves multi-view clustering with superior generalization and robustness performance over previous methods. In future work, we will extend this method to deal with other tasks.

\bibliographystyle{named}

\begin{thebibliography}{}

\bibitem[\protect\citeauthoryear{Alemi \bgroup \em et al.\egroup
  }{2017}]{alemi2017deep}
Alexander~A. Alemi, Ian Fischer, Joshua~V. Dillon, and Kevin Murphy.
\newblock Deep variational information bottleneck.
\newblock In {\em International Conference on Learning Representations}, 2017.

\bibitem[\protect\citeauthoryear{Cai \bgroup \em et al.\egroup
  }{2012}]{bioinformatics}
Xiao Cai, Hua Wang, Heng Huang, and Chris Ding.
\newblock {Joint stage recognition and anatomical annotation of drosophila gene
  expression patterns}.
\newblock {\em Bioinformatics}, 28(12):i16--i24, 2012.

\bibitem[\protect\citeauthoryear{Chen \bgroup \em et al.\egroup
  }{2020}]{chen2020simple}
Ting Chen, Simon Kornblith, Mohammad Norouzi, and Geoffrey Hinton.
\newblock A simple framework for contrastive learning of visual
  representations.
\newblock In {\em International conference on machine learning}, pages
  1597--1607. PMLR, 2020.

\bibitem[\protect\citeauthoryear{Federici \bgroup \em et al.\egroup
  }{2020}]{Federici2020Learning}
Marco Federici, Anjan Dutta, Patrick Forré, Nate Kushman, and Zeynep Akata.
\newblock Learning robust representations via multi-view information
  bottleneck.
\newblock In {\em International Conference on Learning Representations}, 2020.

\bibitem[\protect\citeauthoryear{Fei-Fei \bgroup \em et al.\egroup
  }{2007}]{FEIFEI200759}
Li~Fei-Fei, Rob Fergus, and Pietro Perona.
\newblock Learning generative visual models from few training examples: An
  incremental bayesian approach tested on 101 object categories.
\newblock {\em Computer Vision and Image Understanding}, 106(1):59--70, 2007.

\bibitem[\protect\citeauthoryear{Hotelling}{1992}]{Hotelling1992}
Harold Hotelling.
\newblock {\em Relations Between Two Sets of Variates}, pages 162--190.
\newblock 1992.

\bibitem[\protect\citeauthoryear{Huang \bgroup \em et al.\egroup
  }{2019}]{ijcai2019p356}
Zhenyu Huang, Joey~Tianyi Zhou, Xi~Peng, Changqing Zhang, Hongyuan Zhu, and
  Jiancheng Lv.
\newblock Multi-view spectral clustering network.
\newblock In {\em Proceedings of the Twenty-Eighth International Joint
  Conference on Artificial Intelligence, {IJCAI-19}}, pages 2563--2569.
  International Joint Conferences on Artificial Intelligence Organization,
  2019.

\bibitem[\protect\citeauthoryear{Khan \bgroup \em et al.\egroup
  }{2019}]{9170204}
Ghufran~Ahmad Khan, Jie Hu, Tianrui Li, Bassoma Diallo, and Qianqian Huang.
\newblock Weighted multi-view data clustering via joint non-negative matrix
  factorization.
\newblock In {\em 2019 IEEE 14th International Conference on Intelligent
  Systems and Knowledge Engineering (ISKE)}, pages 1159--1165, 2019.

\bibitem[\protect\citeauthoryear{Kingma and Ba}{2014}]{kingma2014adam}
Diederik~P Kingma and Jimmy Ba.
\newblock Adam: A method for stochastic optimization.
\newblock {\em arXiv preprint arXiv:1412.6980}, 2014.

\bibitem[\protect\citeauthoryear{Kingma and Welling}{2014}]{Kingma2014}
Diederik~P. Kingma and Max Welling.
\newblock {Auto-Encoding Variational Bayes}.
\newblock In {\em 2nd International Conference on Learning Representations,
  {ICLR} 2014, Banff, AB, Canada, April 14-16, 2014, Conference Track
  Proceedings}, 2014.

\bibitem[\protect\citeauthoryear{Li \bgroup \em et al.\egroup
  }{2019a}]{iccv2019RMLSLMVC}
Ruihuang Li, Changqing Zhang, Huazhu Fu, Xi~Peng, Joey~Tianyi Zhou, and Qinghua
  Hu.
\newblock Reciprocal multi-layer subspace learning for multi-view clustering.
\newblock In {\em 2019 IEEE/CVF International Conference on Computer Vision
  (ICCV)}, pages 8171--8179, 2019.

\bibitem[\protect\citeauthoryear{Li \bgroup \em et al.\egroup
  }{2019b}]{9010687}
Ruihuang Li, Changqing Zhang, Huazhu Fu, Xi~Peng, Joey~Tianyi Zhou, and Qinghua
  Hu.
\newblock Reciprocal multi-layer subspace learning for multi-view clustering.
\newblock In {\em 2019 IEEE/CVF International Conference on Computer Vision
  (ICCV)}, pages 8171--8179, 2019.

\bibitem[\protect\citeauthoryear{Li \bgroup \em et al.\egroup
  }{2019c}]{ijcai2019p409}
Zhaoyang Li, Qianqian Wang, Zhiqiang Tao, Quanxue Gao, and Zhaohua Yang.
\newblock Deep adversarial multi-view clustering network.
\newblock pages 2952--2958, 2019.

\bibitem[\protect\citeauthoryear{Lin and Kang}{2021}]{ijcaiGFMAGC}
Zhiping Lin and Zhao Kang.
\newblock Graph filter-based multi-view attributed graph clustering.
\newblock In Zhi-Hua Zhou, editor, {\em Proceedings of the Thirtieth
  International Joint Conference on Artificial Intelligence, {IJCAI-21}}, pages
  2723--2729. International Joint Conferences on Artificial Intelligence
  Organization, 2021.

\bibitem[\protect\citeauthoryear{Lin \bgroup \em et al.\egroup
  }{2021a}]{cvpr2021COMPLETER}
Yijie Lin, Yuanbiao Gou, Zitao Liu, Boyun Li, Jiancheng Lv, and Xi~Peng.
\newblock Completer: Incomplete multi-view clustering via contrastive
  prediction.
\newblock In {\em 2021 IEEE/CVF Conference on Computer Vision and Pattern
  Recognition (CVPR)}, pages 11169--11178, 2021.

\bibitem[\protect\citeauthoryear{Lin \bgroup \em et al.\egroup
  }{2021b}]{COMPLETER}
Yijie Lin, Yuanbiao Gou, Zitao Liu, Boyun Li, Jiancheng Lv, and Xi~Peng.
\newblock Completer: Incomplete multi-view clustering via contrastive
  prediction.
\newblock In {\em 2021 IEEE/CVF Conference on Computer Vision and Pattern
  Recognition (CVPR)}, pages 11169--11178, 2021.

\bibitem[\protect\citeauthoryear{Liu \bgroup \em et al.\egroup
  }{2013}]{MVCJNMF}
Jialu Liu, Chi Wang, Jing Gao, and Jiawei Han.
\newblock {\em Multi-View Clustering via Joint Nonnegative Matrix
  Factorization}, pages 252--260.
\newblock 2013.

\bibitem[\protect\citeauthoryear{MacQueen}{1967}]{MacQueen1967SomeMF}
J.~MacQueen.
\newblock Some methods for classification and analysis of multivariate
  observations.
\newblock 1967.

\bibitem[\protect\citeauthoryear{Nie \bgroup \em et al.\egroup
  }{2017}]{ijcai2017p357}
Feiping Nie, Jing Li, and Xuelong Li.
\newblock Self-weighted multiview clustering with multiple graphs.
\newblock pages 2564--2570, 2017.

\bibitem[\protect\citeauthoryear{Peng \bgroup \em et al.\egroup
  }{2019}]{pmlr-v97-peng19a}
Xi~Peng, Zhenyu Huang, Jiancheng Lv, Hongyuan Zhu, and Joey~Tianyi Zhou.
\newblock {COMIC}: Multi-view clustering without parameter selection.
\newblock In Kamalika Chaudhuri and Ruslan Salakhutdinov, editors, {\em
  Proceedings of the 36th International Conference on Machine Learning},
  volume~97, pages 5092--5101. PMLR, 2019.

\bibitem[\protect\citeauthoryear{Van~der Maaten and
  Hinton}{2008}]{van2008visualizing}
Laurens Van~der Maaten and Geoffrey Hinton.
\newblock Visualizing data using t-sne.
\newblock {\em Journal of machine learning research}, 9(11), 2008.

\bibitem[\protect\citeauthoryear{Wan \bgroup \em et al.\egroup
  }{2021}]{wan2021multi}
Zhibin Wan, Changqing Zhang, Pengfei Zhu, and Qinghua Hu.
\newblock Multi-view information-bottleneck representation learning.
\newblock In {\em Proceedings of the AAAI Conference on Artificial
  Intelligence}, volume~35, 2021.

\bibitem[\protect\citeauthoryear{Wang \bgroup \em et al.\egroup
  }{2019}]{wang2019deep}
Qi~Wang, Claire Boudreau, Qixing Luo, Pang-Ning Tan, and Jiayu Zhou.
\newblock Deep multi-view information bottleneck.
\newblock In {\em Proceedings of the 2019 SIAM International Conference on Data
  Mining}, pages 37--45. SIAM, 2019.

\bibitem[\protect\citeauthoryear{Xu \bgroup \em et al.\egroup
  }{2021}]{DEMVCINS2021}
Jie Xu, Yazhou Ren, Guofeng Li, Lili Pan, Ce~Zhu, and Zenglin Xu.
\newblock Deep embedded multi-view clustering with collaborative training.
\newblock {\em Information Sciences}, 573:279--290, 2021.

\bibitem[\protect\citeauthoryear{Xu \bgroup \em et al.\egroup }{2022a}]{SDMVC}
Jie Xu, Yazhou Ren, Huayi Tang, Zhimeng Yang, Lili Pan, Yang Yang, Xiaorong Pu,
  Philip~S. Yu, and Lifang He.
\newblock Self-supervised discriminative feature learning for deep multi-view
  clustering.
\newblock {\em IEEE Transactions on Knowledge and Data Engineering}, pages
  1--12, 2022.

\bibitem[\protect\citeauthoryear{Xu \bgroup \em et al.\egroup }{2022b}]{MFLVC}
Jie Xu, Huayi Tang, Yazhou Ren, Liang Peng, Xiaofeng Zhu, and Lifang He.
\newblock Multi-level feature learning for contrastive multi-view clustering.
\newblock In {\em 2022 IEEE/CVF Conference on Computer Vision and Pattern
  Recognition (CVPR)}, pages 16030--16039, 2022.

\bibitem[\protect\citeauthoryear{Zhang \bgroup \em et al.\egroup
  }{2019}]{AE2NET}
Changqing Zhang, Yeqing Liu, and Huazhu Fu.
\newblock Ae2-nets: Autoencoder in autoencoder networks.
\newblock In {\em 2019 IEEE/CVF Conference on Computer Vision and Pattern
  Recognition (CVPR)}, pages 2572--2580, 2019.

\bibitem[\protect\citeauthoryear{Zheng \bgroup \em et al.\egroup
  }{2021}]{CUMRLTCSVT2021}
Qinghai Zheng, Jihua Zhu, and Zhongyu Li.
\newblock Collaborative unsupervised multi-view representation learning.
\newblock {\em IEEE Transactions on Circuits and Systems for Video Technology},
  32(7):4202--4210, 2021.

\bibitem[\protect\citeauthoryear{Zheng \bgroup \em et al.\egroup
  }{2022}]{GUMRL}
Qinghai Zheng, Jihua Zhu, Zhongyu Li, and Haoyu Tang.
\newblock Graph-guided unsupervised multi-view representation learning.
\newblock {\em IEEE Transactions on Circuits and Systems for Video Technology},
  2022.

\bibitem[\protect\citeauthoryear{Zhou and Shen}{2020}]{EtEAANMMC}
Runwu Zhou and Yi-Dong Shen.
\newblock End-to-end adversarial-attention network for multi-modal clustering.
\newblock In {\em 2020 IEEE/CVF Conference on Computer Vision and Pattern
  Recognition (CVPR)}, pages 14607--14616, 2020.

\end{thebibliography}

\end{document}